\def\year{2017}\relax
\documentclass[letterpaper]{article} 
\usepackage{aaai17}  
\usepackage{times}  
\usepackage{helvet}  
\usepackage{courier}  
\usepackage{url}  
\usepackage{graphicx}  

\begin{document}
%
\title{Representational Issues in the Debate on the Standard Model of the Mind}
\author{Antonio Chella* Marcello Frixione** Antonio Lieto***\\
**University of Palermo and ICAR-CNR, Italy\\
***University of Genoa, Italy\\
****University of Turin and ICAR-CNR, Italy\\
antonio.chella@unipa.it, marcello.frixione@unige.it, lieto@di.unito.it}


\maketitle
\begin{abstract}
In this paper we discuss some of the issues concerning the \emph{Memory and Content} aspects in the recent debate on the identification of a Standard  Model of the Mind \cite{laird2017}. In particular we focus on the representational models concerning the Declarative Memories of current Cognitive Architectures (CAs). In doing so we outline some of the main problems affecting the current CAs and suggest that the Conceptual Spaces, a representational framework developed by G\"ardenfors, is worth-considering to address such problems. Finally we briefly analyze the alternative representational assumptions employed in the three CAs constituting the current baseline for the Standard Model (i.e. SOAR, ACT-R and Sigma).  In doing so, we point out the respective differences and discuss their implications in the light of the analyzed problems. 
 
\end{abstract}





\section{Introduction}
In the last decades, many Cognitive Architectures (CAs) have been realized adopting different assumptions about the organization and the representation of their knowledge level. Some of them adopt a symbolic approach, some are based on a purely connectionist model, while others  adopt a hybrid approach combining connectionist and symbolic representational levels. In this paper we suggest that, among the different approaches that are worth-considering in the debate concerning the identification of a standard representational model in artificial minds, the framework of Conceptual Spaces can play an important role for connecting symbolic, subsymbolic and diagrammatic representations and for dealing with some problematic aspects affecting the knowledge level in CAs. Finally, we analyze the alternative representational models that have been employed in the three CAs constituting the baseline of the current Standard Model, by pointing out the respective differences and their implications.

\subsection{Representational Limits of current CAs}

It has been recently argued that two of the main current limitations of the knowledge level of the CAs are represented by the limited \emph{size}  and the \emph{homogeneous typology} of the encoded and processed knowledge \cite{lieto2017knowledge}. While the size problem corresponds to the fact that CAs usually operate with very limited and ad-hoc built knowledge bases, the problem concerning the homogeneity issue concerns the fact that usually the type of knowledge represented and manipulated by most CAs (including those provided with extended knowledge modules) mainly covers the so called classical part of conceptual information (i.e. that one representing concepts in terms of necessary and sufficient conditions, see \cite{frixione2012representing} on these aspects). On the other hand, the so called €œcommon-sense knowledge components (i.e. those that, based on the results from the cognitive science, allow to characterize concepts in terms of €œprototypes€, €œexemplars€ or €œtheories€, see \cite{murphy2002big}) is largely absent in such computational frameworks\footnote{There are, however, some proposals explicitly suggesting  to deal with this problem by assuming a heterogeneous representational stance \cite{lieto2014computational}. According to the heterogeneous approach a given concept is endowed by different types of potentially co-existing representations (e.g. prototypes, exemplars etc.). In addition, to each type of representation is associated a corresponding reasoning mechanisms. Handling the interaction between all these reasoning mechanisms represents a crucial aspect of the heterogeneous proposal. In the rest of the paper we assume the heterogeneous representational stance as a way to deal with the knowledge homogeneity problem.}.

As a consequence of this representational aspect, these systems have also a limited capacity of handling, in an integrated way, the heterogeneous amount of co-existing common-sense reasoning mechanisms which are, on the other hand, well established in the psychological literature. Such mechanisms  belongs to the class of typicality-based reasoning (and includes, for example, prototype and exemplar-based categorization).\footnote{A \emph{prototype-based categorization} is obtained, for example, when a stimulus with the following features: ``it has fur, woofs and wags its tail'' is categorized ad a DOG, since these  cues are associated to the prototype of dog. Prototype-based reasoning, however, is not the only type of reasoning based on typicality. In fact, if an exemplar corresponding to the stimulus being categorized is available, too, it is acknowledged that humans use to classify it by evaluating its similarity w.r.t. the exemplar, rather than w.r.t. the prototype associated to the underlying concepts. This type of common sense categorization is known in literature as \emph{exemplars-based categorization} (and the phenomenon according to which the exemplar is favoured with respect to the prototype is known as \emph{old-item advantage effect}). See ~\cite{frixione2013representing} on this aspect.} 


\section{An Intermediate Conceptual Spaces Level}
In our opinion, a possible way to deal with the above mentioned problems can be provided by the adoption of Conceptual Space representations (integrated with other representational formalisms).
Conceptual Spaces \cite{gardenfors00conceptual} have been proposed as an intermediate level of representation between the subsymbolic and symbolic levels. It has been argued that the integration of this level enables to overcome some classical problems specifically related to the subsymbolic and symbolic representations considered in isolation \cite{gardenfors00conceptual}. Conceptual Spaces are a geometrical framework for the representation of knowledge\footnote{In the last fifteen years, such framework has been employed in a vast range of AI applications spanning from visual perception \cite{chella1997cognitive} to robotics  \cite{chella2003anchoring}, question answering \cite{lieto2015common} etc. See \cite{zenker2015applications} for a recent overview.} and can be thought as a \emph{metric} space in which entities are characterized by quality dimensions \cite{gardenfors00conceptual}. To each quality dimension is associated a geometrical (topological or metrical) structure.  In some cases, such dimensions can be directly related to perceptual mechanisms; examples of this kind are temperature, weight, brightness, pitch. In other cases, dimensions can be more abstract in nature. 
In this setting, concepts correspond to convex regions, and regions with different geometrical properties correspond to different sorts of concepts \cite{gardenfors00conceptual}. Here, prototypes and prototypical reasoning have a natural geometrical interpretation: prototypes correspond to the geometrical center of a convex region (the centroid). Also exemplar-based representation can be represented as points in a multidimensional space, and their similarity can be computed as the intervening distance between two points, based on some suitable metrics (such as Euclidean and Manhattan distance etc.). 

Recently some available conceptual categorization systems, explicitly assuming the heterogeneous representational hypothesis and coupling Conceptual Spaces representations and symbolic knowledge bases (i.e. ontologies), have been developed. The DUAL PECCS system \cite{lieto2017dual}, for example, relies on both such assumptions. Such system has been integrated with the Declarative Memories and the knowledge processing mechanisms of different CAs (ACT-R and CLARION). In DUAL PECCS, the interaction of the common-sense categorization strategies (based on prototypes and exemplars representation and operating on Conceptual Spaces representations) and classical deductive categorization mechanisms (executed on the ontological representations) is explicitly provided. 
This aspect is of particular interest in light of the problem concerning the homogeneity of the encoded knowledge. In fact, since the design of the interaction of the different processes operating with heterogeneous representations still represents a largely ignored problem in current CAs, this system shows that Conceptual Spaces represent a relatively effortless framework to both i) model the dynamics between prototype and exemplar-based processes and ii) connect such mechanisms with processes operating on different types of representational formalisms (i.e. the symbolic or logic-oriented ones) that are more suitable to represent conceptual information in classical terms. 

Concerning the size problem, the possible grounding of the Conceptual Spaces representational framework with symbolic structures enables their integration with wide-coverage knowledge bases such CYC (as provided, for example, in DUAL PECCS \cite{lieto2017dual}), DBpedia or similar. Additionally, there are also some initial attempts to automatically learn and encode wide-coverage Conceptual Spaces knowledge bases \cite{derrac2015inducing} also starting by wide-coverage linguistic resources such as BabelNet (\url{http://babelnet.org/}) and ConceptNet (\url{http://conceptnet5.media.mit.edu/}), see \cite{lieto2016resource}. Despite the recent progresses in this sense, however, we acknowledge that there is still a gap to cover in order to produce knowledge bases encoded in terms of Conceptual Spaces that can be comparable with the sizes of the wide-coverage ontological Knowledge Bases (KBs) mentioned above. In principle, however, this framework seems suitable to deal with both the size and the knowledge homogeneity issues.

An additional element of interest concerning the advantages provided by introducing the Conceptual Spaces as an intermediate representational level in CAs regards its capability to address a classical problem in conceptualization: namely the problem of reconciling compositionally and typicality effects \footnote{Broadly speaking this aspect regards the problem of dealing, in a coherent way, with the compositionality of prototypical representations. According to a well-known argument \cite{fodor1981present}, prototypes are not compositional. In brief, the argument runs as follows: consider a concept like \emph{pet fish}. It results from the composition of the concept \emph{pet} and the concept \emph{fish}. However, the prototype of \emph{pet fish} cannot result from the composition of the prototypes of a pet and a fish: a typical pet is furry and warm, a typical fish is grayish, but a typical pet fish is neither furry and warm nor grayish. The possibility of explaining, in a coherent way, this type of combinatorial and generative phenomenon highlights a crucial aspect of the conceptual processing capabilities in human cognition and concerns some crucial high-level cognitive abilities such as that ones concerning conceptual composition, metaphor generation and creative thinking. Dealing with this problem requires the harmonization of two conflicting requirements in representational systems: the need of syntactic, generative, compositionality (typical of logical and symbolic-oriented systems) and that one concerning the exhibition of typicality effects.}.
This aspect does not affect, per se, the size problem but the problem concerning the knowledge homogeneity (since it assumes the existence of typicality-based representations). Such aspect has been shown to be problematic for both symbolic/logic-oriented approaches \cite{osherson1981adequacy}) and for classical connectionist approaches \cite{fodor1988connectionism}. On the other hand, this aspect can be formally handled by recurring to Conceptual Spaces (as shown in \cite{lieto2017conceptual,lewis2016hierarchical}.). In the next sections we briefly outline some arguments additionally supporting the adoption of a Conceptual Spaces representational level in CAs.

\subsection{Interpretation of Neural Networks} 

A relevant issue suggesting the adoption of Conceptual Space in CAs is represented by the possibility of using this representational layer as an interpreter of underlying opaque artificial neural networks (ANN) representations, that, on the other hand, are very well suited for tasks concerning perceptual abilities and are widely used in current CAs. We claim that the theory of Conceptual Spaces can be considered as a sort of designing style that helps to model more transparent neural networks, and it can facilitate the grounding and the interpretation of the hidden layers of units. As a consequence, the interpretation of neural network representations in terms of Conceptual Spaces provides a more abstract and transparent view on the underlying behavior of the networks.

G\"ardenfors \cite{gardenfors00conceptual} offers a simple analysis of the relationship between Conceptual Spaces and Self Organising Maps. Hereafter, \cite{balkenius2016spaces} propose a more articulate interpretation of the widely adopted RBF networks in terms of dimensions of a suitable Conceptual Space 
According to this approach, a neural network built by a set of RBF units can be interpreted as a simple Conceptual Space described by a set of integral quality dimensions. Consequently, a neural network built by a set of sets of RBF units may be geometrically interpreted by a conceptual space made up by sets of integral dimensions.

Additionally, following the Chorus of Prototypes approach proposed by Edelman \cite{edelman1995representation}, the units of an RBF network can be interpreted as prototypes in a suitable Conceptual Space. This interpretation enables the measurement of similarity between the input of the network and the prototypes corresponding to the units. Such an interpretation would have been much more problematic by considering the neural network alone, since this information would have been implicit and hidden. Moreover, it is possible to take into account, for example, the delicate cases of Chimeric entities, which are almost equidistant between two or more prototypes (i.e. the lion and the goat) (see  \cite{edelman1995representation}). This aspect is related to the PET FISH example described in the previous section (footnote 4). In this respect, the capability of accounting for the compositionally based on typicality traits seems to be a crucial feature of the Conceptual Spaces empowering both symbolic and sub-symbolic representations \footnote{It is worth-noting that also some forms of neuro-symbolic integration currently developed in CAs like ACT-R, and belonging to the class of the \emph{neo-connectionist} approaches, allows to deal with the the above mentioned problem by providing a series of mechanisms that are able to deal with limited forms of compositionality in neural networks  \cite{o2013limited} and that can be integrated with additional processes allowing the compatibility with typicality effects. In this respect, such approaches play an equivalent role w.r.t that one played by the Conceptual Spaces on these issues. In addition, however, we claim that Conceptual Spaces can offer a unifying framework for interpreting many kinds of diagrammatic and analogical representations (see the next section). On these classes of representations, limited work has been done by these hybrid neuro-symbolic systems (including ACT-R). This is a symptom that the treatment of their representational and reasoning mechanisms is not trivial in these environments and that often they need to be integrated with external diagrammatic representation systems, see \cite{lieto2017knowledge}.}. An additional element of interests come from the research in computational neuroscience. According to a recent study \cite{reimann2017cliques}, the brain processes information involving cliques of neurons bound into cavities and reacts to external stimuli by building increasingly complex and multidimensional representations starting with rods (1D), then planks (2D), then cubes (3D), and then more complex geometries with 4D, 5D, etc. While the intuitive connection of this finding with the Conceptual Spaces framework is quite evident, this anoalogy in our opinion, deserves further attention. 










\subsection{Unifying Picture-Like Representations} 

Many pictorial, analog or diagrammatic models have been proposed in various fields of Cognitive Science, which take advantage of forms of representations that are picture-like, in the sense that they spatially \emph{resemble} to what they represent \cite{glasgow1995diagrammatic}.




This class of representations is heterogeneous, and it is surely not majoritarian if compared to the main streams of symbolic/logic based systems and of neural networks. Moreover, they lack a general theory, and, despite their intuitive appeal, a common and well understood theoretical framework does not exist.  

Conceptual Spaces, thanks to their geometrical nature, allow the representation of this sort of information and offer, at the same time a general, well understood and theoretically grounded framework that could enable to encompass most of the existing diagrammatic representations. 

The geometrical nature of conceptual spaces can be useful also in representing more abstract and non-specifically spatial domains and phenomena. A typical problem of both symbolic and neural representations regards the ability to track the identity of individual entities over time. The properties of an entity change across the time. At which condition can we re-identify an entity as the same, despite its changes? In many cases the answer is not easy. Conceptual Spaces suggest a way to face the problem. We said that individual objects are represented by points in Conceptual Spaces. However, in a dynamic
perspective, objects can be rather seen as trajectories in a suitable Conceptual Space indexed by time, since the properties
of objects usually change with time. Objects may move, may age, an
object can alter its shape or color, and so on. As the properties
of an object are modified, the point, representing it in
the Conceptual Space, moves according to a certain trajectory. Since usually this modifications happens smoothly and not abruptly, 
several assumptions can be made on this trajectory,
e.g., smoothness, and obedience to physical laws \cite{chella2004perceptual}. 

Figuring out the evolution of an object as its future position,
or the way in which its features are going to change, can
be seen as the extrapolation of a trajectory in a Conceptual
Space. To identify again an object that has been occluded
for a certain time interval amounts to interpolate its past and
present trajectories. In general, this characteristic represents a powerful heuristic to track the identity of an individual object. Also in this case, crucial aspects of diagrammatic representations find a more general and unifying interpretation regarding Conceptual Spaces. 

In the next sections we provide a brief overview of the representational hypotheses adopted by the three different CAs constituting the current baseline for the Standard Model of Mind. In doing so we try to analyze to what extent such systems deal with the problematic aspects discussed above.

\subsection{SOAR}

SOAR was considered by Newell a candidate for a Unified Theory of Cognition \cite{newell1994unified}. In such architecture, all the cognitive tasks can be represented by problem spaces that are searched by production rules grouped into operators. These production rules are fired in parallel to produce reasoning cycles. From a representational perspective, SOAR exploits symbolic representations of knowledge (called chunks) in its the declarative memory (called Semantic Memory) and use pattern matching, and in the more recent versions also spreading activation mechanisms \cite{jones2016efficient}, to select relevant knowledge elements. 


With respect to the knowledge homogeneity issue, the main problem of this architecture relies on the fact  that it does not specify how the typical knowledge components of a concept (that can eventually be represented by adopting a frame-like structure) and the corresponding non monotonic-reasoning strategy can interact with possibly conflicting representational and reasoning procedures characterizing other conceptualisation of the same conceptual entity \footnote{Let us think to the case of WHALE. A prototypical conceptualization would classify whales as a FISH (since a whale share many typical traits with fishes). On the other hand, a classical conceptualization would classify a whale as a MAMMAL.}. In short it assumes, like most of the symbolic-oriented CAs, the availability of a monolithic conceptual structure (e.g., a frame-like prototype or a ``classical" concept) without specifying how such information can be integrated and harmonized with other knowledge components to form the whole knowledge spectrum characterizing a given concept. Therefore the current version of the system is not able to deal, in an integrated perspective, with prototype and exemplar-based categorization. With respect to to the size problem, the SOAR knowledge level is also problematic. SOAR agents, in fact, are not endowed with general knowledge. This problem is acknowledged in \cite{laird2012soar} but there is no available literature attesting progress in this respect \footnote{There are, however, attempts to extend in a efficient way the Semantic Memory of SOAR with external lexical resources such as, for example, Wordnet \cite{derbinsky2010towards}.}. With respect to the diagrammatic representations, finally, SOAR is equipped with a visual imagery module. 


\subsection{ACT-R}
ACT-R \cite{anderson2004integrated} is a cognitive architecture explicitly inspired by theories and experimental results coming from human cognition. Here the cognitive mechanisms concerning the knowledge level emerge from the interaction of two types of knowledge: declarative knowledge, which encodes explicit facts that the system knows, and procedural knowledge, which encodes rules for processing declarative knowledge. In particular, the declarative module is used to store and retrieve pieces of information (called chunks, composed of a type and a set of attribute-value pairs, similar to frame slots) in declarative memory.  ACT-R employs a subsymbolic activation of symbolic conceptual chunks representing the encoded knowledge. Finally, the central production system connects these modules by using a set of IF-THEN production rules. 


Differently from SOAR, ACT-R allows to represent the information in terms of prototypes and exemplars and allow to perform, selectively, either prototype or exemplar-based categorization. This means that the architecture allows the modeller to manually specify which kind of categorization strategy to employ according to his specific needs. Such an architecture, however, only partially addresses the homogeneity problem since it does not allow to represent, jointly, these different types of common-sense representations conveying different types of information for the same conceptual entity (i.e. it does not assume a heterogeneous perspective). As a consequence, it is also not able to autonomously decide which of the corresponding reasoning procedures to activate (e.g. prototypes or exemplars) and to provide a framework able to manage the interaction of such different reasoning strategies (however its overall architectural environment provides, at least in principle, the possibility of implementing cascade reasoning processes triggering one another). 

Even if some attempts exist concerning the design of harmonization strategies between different types of common-sense conceptual categorizations (e.g. exemplar-based and rule-based, see \cite{anderson2001hybrid}) however they do not handle the problem concerning the interaction of the prototype or exemplar-based processes according to the results coming from experimental cognitive science (for example: the old item effect, privileging exemplars w.r.t. prototypes is not modeled. See footnote 2 on this aspect.). Summing up: w.r.t. the homogeneity problem, the components needed to fully reconcile the Heterogeneity approach with ACT-R are available, however they have not been fully exploited yet. 

Regarding the size problem: as for SOAR, ACT-R agents are usually equipped with task-specific knowledge and not with general cross-domain knowledge. In this respect some relevant attempts to overcome this limitation have been recently done by extending the Declarative Memory of the architecture. They will be discussed below along with their current implications. 

\subsection{Sigma}

Sigma is a novel cognitive architecture that starts with the same basic assumption of SOAR \cite{rosenbloom2009towards} and that blends lessons from ACT-R and SOAR with what has been learned separately about graphical models  \cite{laird2017}.

In Sigma the long term memory, as well as  the working memory and perceptual and motor components is grounded in graphical models and, in particular, in factor graphs (a particular type of very efficient graphical models). 

In general, the graphical models can be considered as a class of symbolic representations, where the relations between concepts are weighted by their strength, calculated through statistical
computations. Within the symbolic AI tradition, these models can be seen an attempt to mitigate, for example, the problems concerning common-sense knowledge reasoning\footnote{It is also worth-noting, however, despite the success of the recent statistical approaches in reproducing many cognitive phenomena, that many forms of common-sense knowledge in human cognition do not require predictions about what will happen or, in general, to reason
probabilistically \cite{sloman2014can}. It would be interesting to investigate how such architecture manages these cases.}. 
With respect to the size and the homogeneity issues the current version of the architecture (being also quite new w.r.t the others) seems to encounter problems for both the aspects. At the best of our knowledge, in fact, currently Sigma is not equipped for being extended or integrated with large scale general knowledge bases. With respect to the heterogeneity issues, on the other hand, it allows - in principle - to model forms of approximate reasoning in an efficient way due to the underlying graphical model used as a representational basis. Currently, however, it is not equipped for dealing with, in an integrated way, typicality based reasoning (combining prototypes and exemplars) with standard, "classical", reasoning mechanisms. An interesting aspect, that in our opinion, would be interesting to investigate is to what extent the representational assumption used in Sigma allows align this framework with another well known-theory about the typicality of conceptual knowledge and that is known as \emph{theory-theory} 
\footnote{Theory-theory approaches (see \cite{murphy2002big} for details) explain the typicality effects by assuming that concepts consist of more or less complex mental structures representing (among other things) causal and explanatory relations. Common-sense concepts are mostly characterized in terms of \emph{theories} which are based on arbitrary, i.e. experience-based, rules.}. As for SOAR, finally, also SIGMA supports picture-like visual imagery representations.

\section{Extended Declarative Memories}

Some initial efforts to deal with the size problem have been done (notably all these efforts have been done with ACT-R). To this class of works belongs that one proposed by \cite{oltramari2012pursuing}, aiming at extending the knowledge layer of ACT-R with external ontological content related to the event modelling; that one by  Salvucci \cite{salvucci14endowing}, aiming at enriching the knowledge model of the Declarative Memory (DM) of ACT-R with a world-level knowledge base such as DBpedia (i.e. the semantic version of Wikipedia represented in terms of ontological formalisms), and that one proposed in \cite{ball2004integrating} presenting an integration of the ACT-R Declarative and Procedural Memory with the Cyc ontology \url{http://www.opencyc.org/} (one of the widest ontological resources currently available containing more than 230,000 concepts). The main problematic aspect concerning the extension of the DM with such wide-coverage integrated ontological resources, however, is in that the underlying formalisms of the ontological semantics are mainly biased towards the representation of conceptual information in classical terms (for a more detailed discussion we remind to \cite{lieto2017knowledge}. Other attempts, aimed at extending the DM with knowledge systems able to perform forms of common sense reasoning (such as in in the integration of ACR-R with the SCONE Knowledge Base, see \cite{oltramari2012pursuing}), encounter different problematic issues. For example: with respect to the extensions provided with wide-coverage KBs, the latter approach needs to face the problem concerning the size aspect (since KBs such as SCONE are not comparable in size with Cyc or DBpedia). Concerning the homogeneity problem, on the other hand, such integration seems to provide a straightforward way to combine common-sense reasoning operating with frame-like symbolic knowledge structures. Still, however, the problem concerning the integration of heterogeneous processes acting on different bodies of knowledge is not currently addressed.

In the light of the arguments briefly presented above, it can be argued that the current proposed solutions for dealing with the size and the homogeneity knowledge problems in CAs are not completely satisfactory. In particular, the integrations with huge world-level ontological knowledge bases can be considered a necessary solution for solving the size problem. It is, however, insufficient for dealing with the knowledge homogeneity issue and with the integration of the common-sense conceptual mechanisms that, as assumed in the heterogeneous representational perspective, are activated on heterogeneous bodies of knowledge

\section{Conclusion}
We have provided some arguments supporting the idea that Conceptual Spaces can be a powerful representational framework for dealing with some problematic aspects affecting the knowledge level in the current Long-Term Memories of the CAs (i.e. the size and the knowledge homogeneity problem).  We have also sketched the advantages that such framework, used in combination with symbolic, connectionist and diagrmmatic representations, may provide in general CAs. In our opinion, such evidences support our claim that any standard representational model of mind should be equipped with a geometrical representational level \emph{\`a la} Conceptual Spaces. In the final part of the paper we have proposed an analysis of the representational level of the three CAs currently considered for the development of a Standard Model of the Mind. The analysis shows that, given the size and the knowledge homogeneity problems, the current state of the art is not completely satisfactory and could therefore benefit from the adoption of Conceptual Spaces.

\bibliography{mybibfile}

\begin{thebibliography}{}

\bibitem[\protect\citeauthoryear{Anderson and Betz}{2001}]{anderson2001hybrid}
Anderson, J.~R., and Betz, J.
\newblock 2001.
\newblock A hybrid model of categorization.
\newblock {\em Psychonomic Bulletin \& Review} 8(4):629--647.

\bibitem[\protect\citeauthoryear{Anderson \bgroup et al\mbox.\egroup
  }{2004}]{anderson2004integrated}
Anderson, J.~R.; Bothell, D.; Byrne, M.~D.; Douglass, S.; Lebiere, C.; and Qin,
  Y.
\newblock 2004.
\newblock An integrated theory of the mind.
\newblock {\em Psychological review} 111(4):1036.

\bibitem[\protect\citeauthoryear{Balkenius and
  G{\"a}rdenfors}{2016}]{balkenius2016spaces}
Balkenius, C., and G{\"a}rdenfors, P.
\newblock 2016.
\newblock Spaces in the brain: From neurons to meanings.
\newblock {\em Frontiers in psychology} 7.

\bibitem[\protect\citeauthoryear{Ball, Rodgers, and
  Gluck}{2004}]{ball2004integrating}
Ball, J.; Rodgers, S.; and Gluck, K.
\newblock 2004.
\newblock Integrating act-r and cyc in a large-scale model of language
  comprehension for use in intelligent agents.
\newblock In {\em AAAI workshop},  19--25.

\bibitem[\protect\citeauthoryear{Chella \bgroup et al\mbox.\egroup
  }{2004}]{chella2004perceptual}
Chella, A.; Coradeschi, S.; Frixione, M.; and Saffiotti, A.
\newblock 2004.
\newblock Perceptual anchoring via conceptual spaces.
\newblock In {\em Proceedings of the AAAI-04 Workshop on Anchoring Symbols to
  Sensor Data},  40--45.

\bibitem[\protect\citeauthoryear{Chella, Frixione, and
  Gaglio}{1997}]{chella1997cognitive}
Chella, A.; Frixione, M.; and Gaglio, S.
\newblock 1997.
\newblock A cognitive architecture for artificial vision.
\newblock {\em Artificial Intelligence} 89(1):73--111.

\bibitem[\protect\citeauthoryear{Chella, Frixione, and
  Gaglio}{2003}]{chella2003anchoring}
Chella, A.; Frixione, M.; and Gaglio, S.
\newblock 2003.
\newblock Anchoring symbols to conceptual spaces: the case of dynamic
  scenarios.
\newblock {\em Robotics and Autonomous Systems} 43(2):175--188.

\bibitem[\protect\citeauthoryear{Derbinsky, Laird, and
  Smith}{2010}]{derbinsky2010towards}
Derbinsky, N.; Laird, J.~E.; and Smith, B.
\newblock 2010.
\newblock Towards efficiently supporting large symbolic declarative memories.
\newblock 1001:48109--2121.

\bibitem[\protect\citeauthoryear{Derrac and
  Schockaert}{2015}]{derrac2015inducing}
Derrac, J., and Schockaert, S.
\newblock 2015.
\newblock Inducing semantic relations from conceptual spaces: a data-driven
  approach to plausible reasoning.
\newblock {\em Artificial Intelligence} 228:66--94.

\bibitem[\protect\citeauthoryear{Edelman}{1995}]{edelman1995representation}
Edelman, S.
\newblock 1995.
\newblock Representation, similarity, and the chorus of prototypes.
\newblock {\em Minds and Machines} 5(1):45--68.

\bibitem[\protect\citeauthoryear{Fodor and
  Pylyshyn}{1988}]{fodor1988connectionism}
Fodor, J.~A., and Pylyshyn, Z.~W.
\newblock 1988.
\newblock Connectionism and cognitive architecture: A critical analysis.
\newblock {\em Cognition} 28(1-2):3--71.

\bibitem[\protect\citeauthoryear{Fodor}{1981}]{fodor1981present}
Fodor, J.~A.
\newblock 1981.
\newblock The present status of the innateness controversy.
\newblock In Fodor, J.~A., ed., {\em Representations: Philosophical Essays on
  the Foundations of Cognitive Science}. Cambridge, MA: MIT Press.
\newblock chapter~10,  257 -- 316.

\bibitem[\protect\citeauthoryear{Frixione and
  Lieto}{2012}]{frixione2012representing}
Frixione, M., and Lieto, A.
\newblock 2012.
\newblock Representing concepts in formal ontologies: Compositionality vs.
  typicality effects.
\newblock {\em Logic and Logical Philosophy} 21(4):391--414.

\bibitem[\protect\citeauthoryear{Frixione and
  Lieto}{2013}]{frixione2013representing}
Frixione, M., and Lieto, A.
\newblock 2013.
\newblock Representing non classical concepts in formal ontologies: Prototypes
  and exemplars.
\newblock In {\em New Challenges in Distributed Information Filtering and
  Retrieval}. Springer.
\newblock  171--182.

\bibitem[\protect\citeauthoryear{G{\"a}rdenfors}{2000}]{gardenfors00conceptual}
G{\"a}rdenfors, P.
\newblock 2000.
\newblock {\em Conceptual spaces: The geometry of thought}.
\newblock MIT press.

\bibitem[\protect\citeauthoryear{Glasgow, Narayanan, and
  Chandrasekaran}{1995}]{glasgow1995diagrammatic}
Glasgow, J.; Narayanan, N.~H.; and Chandrasekaran, B.
\newblock 1995.
\newblock {\em Diagrammatic reasoning: Cognitive and computational
  perspectives}.
\newblock Mit Press.

\bibitem[\protect\citeauthoryear{Jones, Wandzel, and
  Laird}{2016}]{jones2016efficient}
Jones, S.~J.; Wandzel, A.~R.; and Laird, J.~E.
\newblock 2016.
\newblock Efficient computation of spreading activation using lazy evaluation.
\newblock {\em Ann Arbor} 1001:48109--2121.

\bibitem[\protect\citeauthoryear{Laird, Lebiere, and Rosenbloom}{in
  press}]{laird2017}
Laird, J.~E.; Lebiere, C.; and Rosenbloom, P.~S.
\newblock in press.
\newblock A standard model of the mind: Toward a common computational framework
  across artificial intelligence, cognitive science, neuroscience, and
  robotics.
\newblock {\em AI Magazine}  1--19.

\bibitem[\protect\citeauthoryear{Laird}{2012}]{laird2012soar}
Laird, J.
\newblock 2012.
\newblock {\em The Soar cognitive architecture}.
\newblock MIT Press.

\bibitem[\protect\citeauthoryear{Lewis and Lawry}{2016}]{lewis2016hierarchical}
Lewis, M., and Lawry, J.
\newblock 2016.
\newblock Hierarchical conceptual spaces for concept combination.
\newblock {\em Artificial Intelligence} 237:204--227.

\bibitem[\protect\citeauthoryear{Lieto, Chella, and
  Frixione}{2017}]{lieto2017conceptual}
Lieto, A.; Chella, A.; and Frixione, M.
\newblock 2017.
\newblock Conceptual spaces for cognitive architectures: A lingua franca for
  different levels of representation.
\newblock {\em Biologically Inspired Cognitive Architectures} 19:1--9.

\bibitem[\protect\citeauthoryear{Lieto, Lebiere, and
  Oltramari}{2017}]{lieto2017knowledge}
Lieto, A.; Lebiere, C.; and Oltramari, A.
\newblock 2017.
\newblock The knowledge level in cognitive architectures: Current limitations
  and possible developments.
\newblock {\em Cognitive Systems Research}.

\bibitem[\protect\citeauthoryear{Lieto, Mensa, and
  Radicioni}{2016}]{lieto2016resource}
Lieto, A.; Mensa, E.; and Radicioni, D.~P.
\newblock 2016.
\newblock A resource-driven approach for anchoring linguistic resources to
  conceptual spaces.
\newblock In {\em AI* IA 2016 Advances in Artificial Intelligence}. Springer.
\newblock  435--449.

\bibitem[\protect\citeauthoryear{Lieto, Radicioni, and
  Rho}{2015}]{lieto2015common}
Lieto, A.; Radicioni, D.~P.; and Rho, V.
\newblock 2015.
\newblock A common-sense conceptual categorization system integrating
  heterogeneous proxytypes and the dual process of reasoning.
\newblock In {\em In Proceedings of the International Joint Conference on
  Artificial Intelligence (IJCAI), Buenos Aires, AAAI Press},  875--881.

\bibitem[\protect\citeauthoryear{Lieto, Radicioni, and
  Rho}{2017}]{lieto2017dual}
Lieto, A.; Radicioni, D.~P.; and Rho, V.
\newblock 2017.
\newblock Dual peccs: a cognitive system for conceptual representation and
  categorization.
\newblock {\em Journal of Experimental \& Theoretical Artificial Intelligence}
  29(2):433--452.

\bibitem[\protect\citeauthoryear{Lieto}{2014}]{lieto2014computational}
Lieto, A.
\newblock 2014.
\newblock A computational framework for concept representation in cognitive
  systems and architectures: Concepts as heterogeneous proxytypes.
\newblock {\em Procedia Computer Science} 41:6--14.

\bibitem[\protect\citeauthoryear{Murphy}{2002}]{murphy2002big}
Murphy, G.~L.
\newblock 2002.
\newblock {\em The big book of concepts}.
\newblock MIT press.

\bibitem[\protect\citeauthoryear{Newell}{1994}]{newell1994unified}
Newell, A.
\newblock 1994.
\newblock {\em Unified theories of cognition}.
\newblock Harvard University Press.

\bibitem[\protect\citeauthoryear{Oltramari and
  Lebiere}{2012}]{oltramari2012pursuing}
Oltramari, A., and Lebiere, C.
\newblock 2012.
\newblock Pursuing artificial general intelligence by leveraging the knowledge
  capabilities of act-r.
\newblock In {\em Artificial General Intelligence}. Springer.
\newblock  199--208.

\bibitem[\protect\citeauthoryear{O'Reilly \bgroup et al\mbox.\egroup
  }{2013}]{o2013limited}
O'Reilly, R.~C.; Petrov, A.~A.; Cohen, J.~D.; Lebiere, C.~J.; Herd, S.~A.;
  Kriete, T.; Calvo, I.~P.; and Symons, J.
\newblock 2013.
\newblock How limited systematicity emerges: A computational cognitive
  neuroscience approach.

\bibitem[\protect\citeauthoryear{Osherson and
  Smith}{1981}]{osherson1981adequacy}
Osherson, D.~N., and Smith, E.~E.
\newblock 1981.
\newblock On the adequacy of prototype theory as a theory of concepts.
\newblock {\em Cognition} 9(1):35--58.

\bibitem[\protect\citeauthoryear{Reimann \bgroup et al\mbox.\egroup
  }{2017}]{reimann2017cliques}
Reimann, M.~W.; Nolte, M.; Scolamiero, M.; Turner, K.; Perin, R.; Chindemi, G.;
  D{\l}otko, P.; Levi, R.; Hess, K.; and Markram, H.
\newblock 2017.
\newblock Cliques of neurons bound into cavities provide a missing link between
  structure and function.
\newblock {\em Frontiers in Computational Neuroscience} 11:48.

\bibitem[\protect\citeauthoryear{Rosenbloom}{2009}]{rosenbloom2009towards}
Rosenbloom, P.~S.
\newblock 2009.
\newblock Towards a new cognitive hourglass: Uniform implementation of
  cognitive architecture via factor graphs.
\newblock In {\em Proceedings of the 9th international conference on cognitive
  modeling},  116--121.

\bibitem[\protect\citeauthoryear{Salvucci}{2014}]{salvucci14endowing}
Salvucci, D.~D.
\newblock 2014.
\newblock Endowing a cognitive architecture with world knowledge.
\newblock In {\em Procs. of the 36th Annual Meeting of the Cognitive Science
  Society}.

\bibitem[\protect\citeauthoryear{Sloman}{2014}]{sloman2014can}
Sloman, A.
\newblock 2014.
\newblock How can we reduce the gulf between artificial and natural
  intelligence?
\newblock In {\em AIC},  1--13.

\bibitem[\protect\citeauthoryear{Zenker and
  G\"{a}rdenfors}{2015}]{zenker2015applications}
Zenker, F., and G\"{a}rdenfors, P.
\newblock 2015.
\newblock {\em Applications of conceptual spaces - The case for geometric
  knowledge representation}.
\newblock Springer.

\end{thebibliography}
\bibliographystyle{aaai}

\end{document}